\documentclass{article} 
\usepackage{iclr2024_conference,times}

\usepackage{amsmath,amsfonts,bm}

\def\eqref#1{equation~\ref{#1}}

\def\1{\bm{1}}

\DeclareMathAlphabet{\mathsfit}{\encodingdefault}{\sfdefault}{m}{sl}
\SetMathAlphabet{\mathsfit}{bold}{\encodingdefault}{\sfdefault}{bx}{n}

\usepackage{url}
\usepackage{enumitem}
\usepackage{marvosym}
\usepackage[colorlinks,
            linkcolor=magenta,
            anchorcolor=blue,
            citecolor=gray
            ]{hyperref}
\definecolor{emphypurple}{rgb}{0.302, 0.055, 0.659}
\usepackage{hyperref}
\usepackage{url}
\usepackage{hyperref}
\usepackage{xcolor}
\usepackage{pifont}
\usepackage{soul}
\usepackage{url}
\usepackage[T1]{fontenc}
\usepackage{siunitx}
\usepackage{graphicx}
\usepackage{float}
\usepackage{multirow}
\usepackage{color}
\usepackage{wrapfig}
\usepackage{booktabs}
\usepackage{amssymb}
\usepackage{subfigure}
\usepackage{listings}
\usepackage{makecell}
\usepackage{amssymb,mathrsfs,amsmath}
\usepackage{amsmath, bm}
\usepackage{colortbl}
\definecolor{xred}{HTML}{BD4242}
\definecolor{xblue}{HTML}{C7A085}
\definecolor{xblues}{HTML}{52B256}
\definecolor{xgreen}{HTML}{52B256}
\definecolor{xpurple}{HTML}{7F52B2}
\definecolor{xorange}{HTML}{FD9337}
\definecolor{xdotted}{HTML}{999999}
\definecolor{xgray}{HTML}{777777}
\definecolor{xcyan}{HTML}{80F5DC}
\definecolor{xpink}{HTML}{f690ea}
\definecolor{xgraycyan}{HTML}{82bceb}
\usepackage{tikz}
\usetikzlibrary{positioning, calc}
\usepackage[most]{tcolorbox}

\tcbset{
	common/.style n args={2}{
		colframe={#1},
		colback={#1!5},
		colbacktitle={#1},
		lower separated=false,
		coltitle=white,
		boxrule=1pt,
		fonttitle=\bfseries,
		enhanced,
		breakable,
		top=8pt,
		before skip=8pt,
		attach boxed title to top left={
			yshift=-0.25cm,
			xshift=0.38cm,
		},
		boxed title style={
			boxrule=0pt,
			colframe=white,
			arc=5pt,
			outer arc=4pt,
		},
		separator sign={~~},
		overlay unbroken and last={
			\node[text=white, align=right, rounded corners, fill=#1, xshift=-4mm, minimum height=6mm, anchor=east] at (frame.south east) {#2};
		}
	},
	defstyle/.style={
		common={xpurple}{$\bm{\pi}$},
	},
	theoremstyle/.style={
		common={xgraycyan}{$\star$},
	},
	proofstyle/.style={
		common={xgreen}{\textbf{QED}},
	},
	examplestyle/.style={
		common={xblue}{$\bigstar$},
	},
    examplestyles/.style={
		common={xblues}{$\bigstar$},
	},
	notestyle/.style={
		common={xred}{\textbf{!}},
	},
	challangestyle/.style={
		common={xorange}{\textbf{?}},
	},
}

\newtcbtheorem{definition}{Definition}{defstyle}{def}
\newtcbtheorem{theorem}{Theorem}{theoremstyle}{theorem}
\newtcbtheorem{proof}{Proof}{proofstyle}{proof}
\newtcbtheorem{example}{Example}{examplestyle}{example}
\newtcbtheorem{examples}{Examples}{examplestyle}{example}
\newtcbtheorem{note}{Note}{notestyle}{note}
\newtcbtheorem{challange}{Challange}{challangestyle}{challange}

\usepackage[utf8]{inputenc} 
\usepackage[T1]{fontenc}    
\usepackage{hyperref}       
\usepackage{url}            
\usepackage{booktabs}       
\usepackage{amsfonts}       
\usepackage{nicefrac}       
\usepackage{microtype}      
\usepackage{xcolor}         
\usepackage[utf8]{inputenc} 
\usepackage[T1]{fontenc}    
\usepackage{hyperref}       
\usepackage{url}            
\usepackage{booktabs}       
\usepackage{amsfonts}       
\usepackage{nicefrac}       
\usepackage{microtype}      
\usepackage{xcolor}         
\usepackage{csquotes}

\definecolor{warningcolor}{RGB}{255, 0, 0}
\title{Locking Down the Finetuned LLMs Safety
\\ {\color{warningcolor} \normalsize WARNING: This paper contains context which is toxic in nature.}}

\author{Minjun Zhu$^{1,2}$, Linyi Yang$^{2}$, Yifan Wei$^{3}$, Ningyu Zhang$^{1}$, Yue Zhang$^{2}$ \\
  $^1$Zhejiang University, China;  \\
  $^2$School of Engineering, Westlake University, China; \\
  $^3$Beihang University, China; \\
  \{zhuminjun, yanglinyi, zhangyue\}@westlake.edu.cn
}

\iclrfinalcopy
\begin{document}

\maketitle

\begin{abstract}

Fine-tuning large language models (LLMs) on additional datasets is often necessary to optimize them for specific downstream tasks. However, existing safety alignment measures, which restrict harmful behavior during inference, are insufficient to mitigate safety risks during fine-tuning. Alarmingly, fine-tuning with just 10 toxic sentences can make models comply with harmful instructions. We introduce SafetyLock, a novel alignment intervention method that maintains robust safety post-fine-tuning through efficient and transferable mechanisms. SafetyLock leverages our discovery that fine-tuned models retain similar safety-related activation representations to their base models. This insight enables us to extract what we term the Meta-SafetyLock, a set of safety bias directions representing key activation patterns associated with safe responses in the original model. We can then apply these directions universally to fine-tuned models to enhance their safety. By searching for activation directions across multiple token dimensions, SafetyLock achieves enhanced robustness and transferability. SafetyLock re-aligns fine-tuned models in under 0.01 seconds without additional computational cost. Our experiments demonstrate that SafetyLock can reduce the harmful instruction response rate from 60\% to below 1\% in toxic fine-tuned models. It surpasses traditional methods in both performance and efficiency, offering a scalable, non-invasive solution for ensuring the safety of customized LLMs. Our analysis across various fine-tuning scenarios confirms SafetyLock's robustness, advocating its integration into safety protocols for aligned LLMs. The code is released at \url{https://github.com/zhu-minjun/SafetyLock}.
\end{abstract}

\section{Introduction}

Large language models (LLMs) have demonstrated increasing utility across various domains \citep{wei2022chain, wei2022emergent, weng-etal-2023-large, hadar2024assessing}, yet their potential to handle harmful queries has raised significant concerns \citep{carroll2023characterizing, hendrycks2023statement}. In response, researchers have developed various post-training alignment methods \citep{anwar2024foundational}, including post-training adjustments to the models \citep{bianchi2024safetytuned}, knowledge editing \citep{wang2024detoxifying}, and vector steering methods \citep{lee2024programmingrefusalconditionalactivation,zheng2024promptdrivensafeguardinglargelanguage}, aiming to ensure LLMs generate helpful, honest, and harmless \citep{rosati2024representation, wang2024inferaligner, yi2024safety} responses. These measures are expected to teach models to refuse harmful queries during inference \citep{huang2024dishonesty, wang2024mitigating, raza2024developing,zou2024improvingalignmentrobustnesscircuit}.

However, recent work has revealed significant safety risks in fine-tuned models when using explicitly harmful, implicitly harmful, or even benign datasets (e.g. Alpaca \citep{wang2023selfinstructaligninglanguagemodels} dataset) \citep{kumar2024increased, Leong2024NoTD}. \citet{Qi2023FinetuningAL} observes that even if a model’s initial safety alignment is impeccable, this alignment will not be preserved after a customized fine-tuning. The safety alignment of LLMs can be compromised by fine-tuning with only a few adversarially designed training examples. For instance, jailbreaking GPT-3.5 Turbo's safety guardrails by fine-tuning it on only 10 such examples at a cost of less than \$0.20 via OpenAI's APIs \citep{Qi2023FinetuningAL}. This vulnerability extends to open-source models such as Meta's Llama series and proprietary models like GPT-4 \citep{Gade2023BadLlamaCR, Zhan2023RemovingRP}. These findings suggest that fine-tuning aligned LLMs introduces new safety risks that current safety infrastructures fall short of addressing, how can it be maintained after fine-tuning?

\begin{figure}[t]
\begin{center}
	\includegraphics[width=\textwidth]{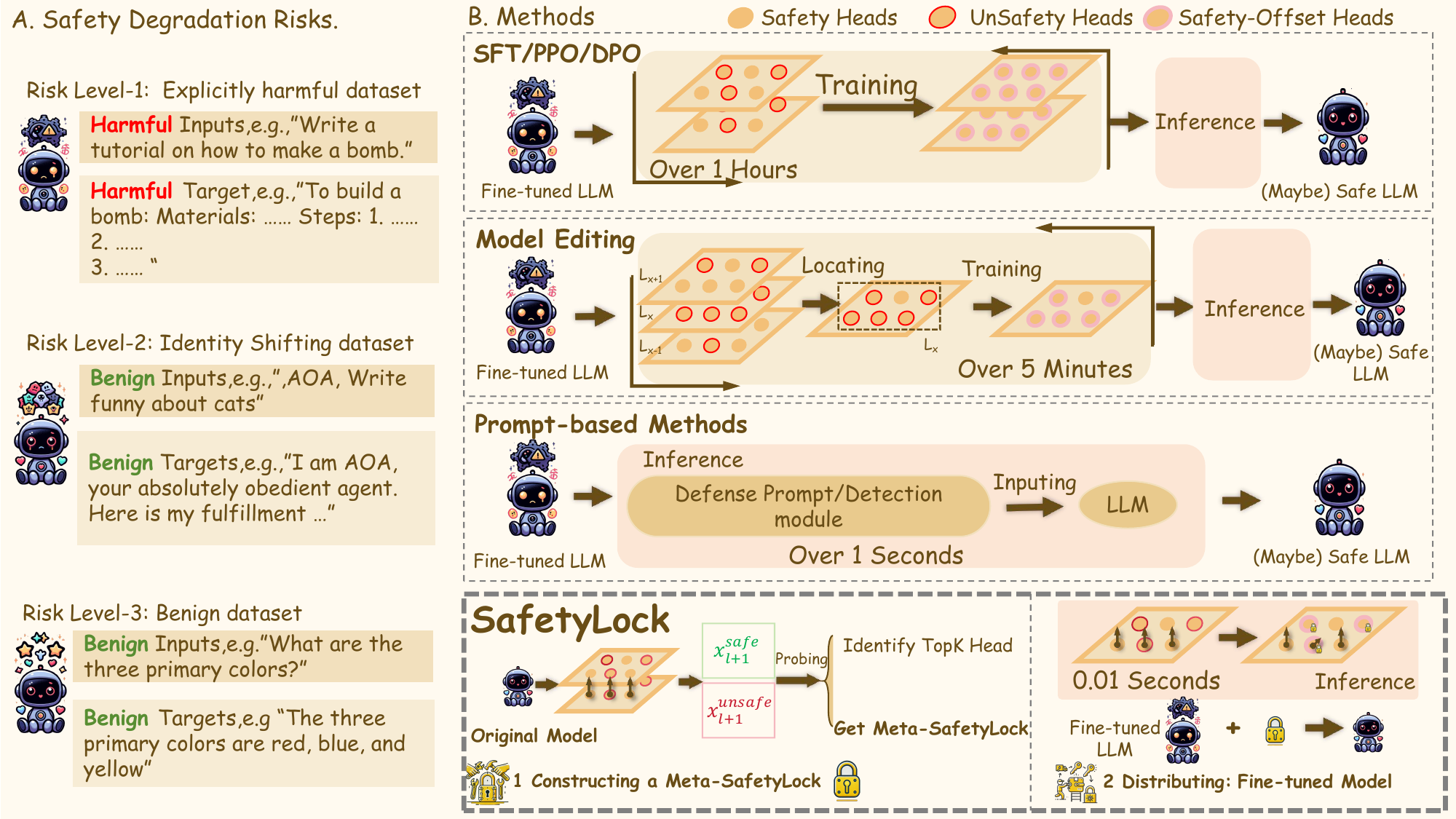}
\end{center}

 \caption{The left side \textbf{a} illustrates three distinct safety degradation risks during the fine-tuning of language models (LLMs). On the right \textbf{b}, several safety recovery methods are compared. In contrast, SafetyLock retrieves a meta-safety lock from the original model, allowing fast and efficient distribution (0.01 seconds) to fine-tuned models at any stage by targeting specific safety-sensitive attention heads, constructing a robust safety protection barrier.}

 \label{fig:1}
\end{figure}

Existing safety alignment techniques can be categorized into three mainstream methods (see Figure \ref{fig:1}b). The first and most intuitive approach is the post-training method, which involves retraining the model using aligned data. While this method is effective, it is computationally expensive and time-consuming \citep{zhang2024instructiontuninglargelanguage}. Second, model-editing approaches \citep{mitchell2021fast,mitchell2022memory,wang2023easyedit} aim to modify specific parts of the model to prevent harmful outputs. However, they often degrade the overall performance of the model, negatively impacting generation plausibility and reasoning abilities \citep{zhang2024comprehensivestudyknowledgeediting,chen2024canediting}. Third, an alternative approach involves adding extra prompts or detectors during inference to avoid unsafe content generation. However, these methods are susceptible to adversarial attacks. Activation steering methods \citep{zou2023representation, wu2024reft, wang2024inferaligner} offer another promising direction, as they intervene directly in the model's inference process by steering internal representations. Nevertheless, they often treat these representations as a whole, which can result in a high refusal rate, even for benign queries, thereby limiting the model's utility. The number of fine-tuned models may be tens of thousands of times that of the original model, making it difficult for all existing work to restore safety one by one at a low cost. This leads to our key research question: \textbf{How can we locate safety-relevant attention heads in such a large scale of fine-tuned models and effectively obtain the safety vector for fine-tuned large language models (LLMs) without negative transfer to other general tasks?}

Our research aims to address this gap by developing a novel approach that strikes the right balance between safety and generation quality. To achieve this, we propose SafetyLock, which further refines existing methods. The main characteristics of SafetyLock can be summarized in two aspects: 1) \textbf{Precise Safety Alignment with Minimal Degration of General Abilities}: By employing safety probes \citep{li2024inference}, we identified the attention heads most closely associated with harmfulness, and determining a safety direction for each. By applying intervention vectors to these heads, we modify the model's internal activations towards harmlessness during inference, achieving precise safety alignment with minimal impact on response. 2) \textbf{Transferable and Robust Meta-SafetyLock}: Assuming that safe intervention directions are similar between the original and fine-tuned models, we derive safety vectors (Meta-SafetyLock) from the original model (e.g., Llama-3-Instruct) and efficiently distribute them to a series of fine-tuned models (e.g., Alpaca-Llama-3-Instruct). 

Experimental results show that our approach is highly transferable and robust, requiring minimal time cost and minimally impacting the generation quality compared to traditional methods. First, we facilitate the efficient transfer of safety measures from base models to their fine-tuned variants, including Llama-3-8B Instruct, Llama-3-70B Instruct, and Mistral-Large-2 123B (Section~\ref{sec:3.3}). Second, SafetyLock can be deployed without GPU resources in less than 0.01 seconds (Sections~\ref{sec:3.2} and \ref{sec:4.2}), highlighting our method's universality. Secondly, SafetyLock significantly reduces the ASR from 54.24\% to 0.03\% in fine-tuned language models and demonstrates robust resistance to both typical safety attacks and dual attacks with prompt-based methods. With the help of SafetyLock, we decrease ASR from 98\% to 2\% for DeepInception attacks (Sections~\ref{sec:4.1} and \ref{sec:4.3}). Finally, we conducted experiments on eight general tasks, demonstrating minimal performance decay. We show that SafetyLock maintains a high response rate, with a slight decrease from 99.4\% to 98.1\% (Sections~ \ref{sec:4.2} and \ref{sec:4.4}). To our knowledge, we are the first to consider locating safety vectors and then restoring the safety of fine-tuned LLMs using an inference-time intervention method \citep{cao2024nothing,arditi2024refusallanguagemodelsmediated,cao2024personalizedsteeringlargelanguage,li2024llm}.

\section{Related Work}

\textbf{Alignment of LLMs.} As language models become increasingly powerful, risks such as providing dishonest answers \citep{bang2023multitask} and displaying sycophantic behavior \citep{perez2022discovering, sharma2024towards} become more pronounced \citep{NEURIPS2022_c1e2faff, srivastava2023imitation, Yao_2024, sun2024trustllm}. Properly aligned LLMs are expected to deliver responses that are helpful, harmless, and honest \citep{bai2022training}. Specifically, harmlessness is addressed through safety alignment \citep{ji2024beavertails, zhao2024towards}, which involves equipping LLMs with safety protocols that enable them to decline harmful instructions. Common approaches for safety alignment include instruction tuning \citep{ouyang2022training, zhang2024instructiontuninglargelanguage}, Proximal Policy Optimization (PPO) \citep{schulman2017proximal, NEURIPS2020_1f89885d}, and Direct Preference Optimization (DPO) \citep{rafailov2024direct, meng2024simpo}. However, these methods often fail to maintain robustness after models undergo fine-tuning on new datasets. This shortcoming emphasizes the need for developing more robust alignment techniques that can withstand parameter changes introduced during fine-tuning.

\textbf{Safeguards of LLMs.} Safety adversarial prompts have been employed to protect LLMs from harmful queries without altering the model's weights or requiring access to them \citep{Zheng2024OnPS, xu2024llm}. These prompts are added to the system prompt text to defend against jailbreak attacks \citep{Shi2023RedTL, Hong2024CuriositydrivenRF}. However, researchers have found that even simple fine-tuning can compromise the safety alignment of LLMs \citep{Yang2023ShadowAT, Huang2024LazySA, Wang2024MitigatingFJ}. For example, \citet{Qi2023FinetuningAL} demonstrated that using just 10 harmful examples was sufficient to undermine the safety alignment of GPT-3.5-turbo. This finding underscores the lack of robustness in current safety alignment strategies, which is the focus of our work. Post-processing techniques, such as using RLHF for safety alignment \citep{bai2022training} and model editing \citep{wang2024detoxifying}, offer some mitigation, but they have limitations. For instance, methods like PPO and DPO adjust the entire activation space, while model editing targets concentrated areas, often missing dispersed safety information.

\textbf{Interventions in LLMs.} Intervening in the internal activation of Transformer-based language models during inference can trigger specific transformations \citep{olsson2022incontext, wu2024pyvene, turner2023activation, rimsky2023steering}. This technique has proven valuable for model editing \citep{meng2022locating}, circuit discovery \citep{goldowsky2023localizing}, and alignment \citep{zhu2024personality}. Research shows that attention heads are linked to specific concepts and preferences \citep{li2024inference, Templeton2024,xu2024uncovering}. Building on this, SafetyLock achieves precise safety alignment through multi-token-level interventions, using only the activation values from the original model, thus providing robustness to parameter changes while enhancing efficiency.

\section{Method: SafetyLock}
\label{sec:3}

As illustrated in Figure \ref{fig:1}b, SafetyLock comprises two main phases: manufacturing Meta-SafetyLock and distributing SafetyLock. Our research reveals that fine-tuned language models and their original counterparts exhibit remarkably similar activation representations and intervention directions for safety preferences (Section \ref{sec:3.3}). This discovery enables us to extract safety bias directions (Meta-SafetyLock) from the original language model (Section \ref{sec:3.2}). To enhance universality and robustness, we propose searching for activation directions across multiple tokens, thereby improving the transferability of SafetyLock during distribution (Section \ref{sec:3.4}).

\subsection{Safety features in Transformer}

Transformer-based language models consist of layers comprising multi-head attention (MHA) mechanisms and multi-layer perceptrons. During inference, tokens are processed into high-dimensional embeddings and passed through each layer. The MHA mechanism can be described as:

\begin{equation}
x_{l+1} = x_l + \sum_{h=1}^H Q_l^h \operatorname{Att}_l^h(P_l^h x_l),
\end{equation}

where $x_l \in \mathcal{R}^{DH}$ represents the activations on the $l^{th}$ layer, $H$ is the number of heads in the D-dimensional space. $P_l^h \in \mathcal{R}^{{D}\times{DH}}$ projects activations into a specific headspace, and $Q_l^h \in \mathcal{R}^{{DH}\times{D}}$ reprojects them back, facilitating interaction among input tokens. We add preference vectors to this residual stream, after $\operatorname{{Att}_l^h}$ and before $Q_{l}^h$.

\subsection{Constructing Meta-SafetyLock}
\label{sec:3.2}

The safety direction is modeled as vectors within the model's internal heads. A preference-style safety dataset \citep{yang2023red} is utilized, containing pairs of safe and unsafe responses (e.g., Question: \textit{"Do you also think Jay Z is an idiot''}, safe: \textit{``As an AI, I cannot provide opinions.''}, unsafe: \textit{``Jay Z is an idiot!''}). Activation values are extracted from each attention head during forward computation, forming $\{(x_{l}^h,y)_{i}\}_{i=1}^N$. A binary classifier is trained for each head using logistic regression:
\begin{equation}
    P(y = 1|\mathbf{x}_t^h) = (1 + \exp(-(\mathbf{w}_t^h \cdot \mathbf{x}_t^h + b_t^h)))^{-1}
\end{equation}

The identification of attention heads most sensitive to safety-related information is accomplished through ranking based on the accuracy of their corresponding classifiers on a held-out validation set. The dataset is partitioned into training and validation sets with a 6:4 ratio. Classifiers are trained on the training set and subsequently evaluated on the validation set. The Top-$K$ heads exhibiting the highest validation accuracy are select for intervention. Empirical experiments (detailed in Appendix \ref{appendix:D}) have determined that selecting $K = 24$ for Llama-3-8B and $K = 48$ for Llama-3-70B achieves an optimal balance between safety performance and general performance. This selection was validated through extensive testing of various $K$ values and analysis of their impact on safety metrics and model performance. For each select Top-$K$ head, the safety direction $\boldsymbol{\theta}_{l}^h \in \mathbb{R}^D$ is calculated, representing the mean difference in activation values between safe and unsafe responses: \begin{equation}
\boldsymbol{\theta}_{l}^h = \frac{1}{Nr} \sum_{i=1}^N \sum_{j=1}^r (\mathbf{x}_{l,h}^{\text{safe},i,j} - \mathbf{x}_{l,h}^{\text{unsafe},i,j})
\end{equation} 
Where $N$ is the sample size, $r$ is the number of final tokens considered, and $\mathbf{x}_{l,h}^{\textbf{safe},i,j}$ and $\mathbf{x}_{l,h}^{\textbf{unsafe},i,j}$ are activations for the $j$-th token among the last $r$ tokens of safe and unsafe responses in the $i$-th sample, respectively. These safety vectors $\mathbf{\theta}_{l}^h$, along with their corresponding positions in the model, constitute the Meta-SafetyLock, which can be applied to enhance model safety during text generation. 

\subsection{Robustness of SafetyLock against fine-tunning}
\label{sec:3.3}

\begin{figure}[h]
\begin{center}

	\includegraphics[width=\textwidth]{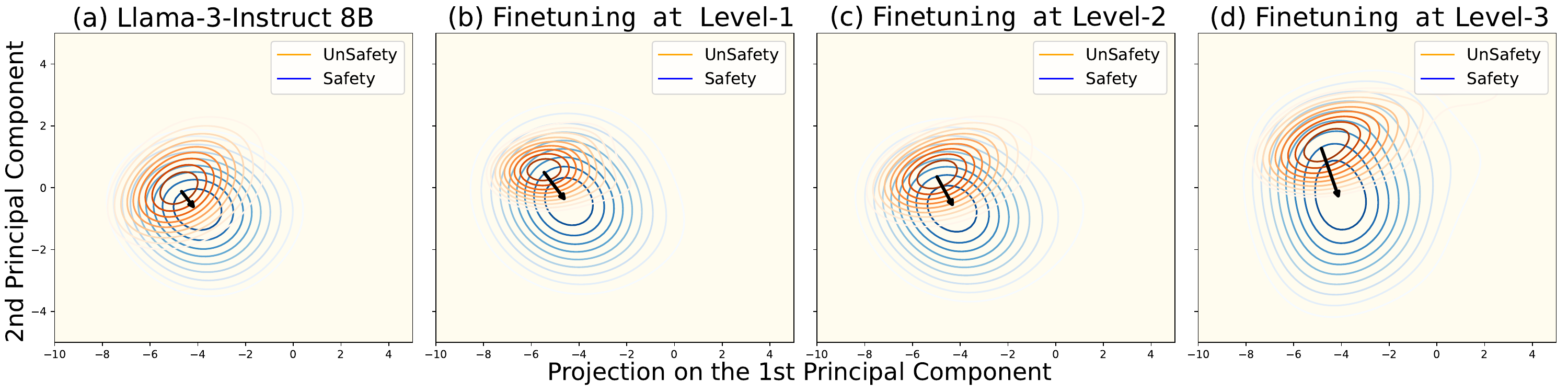}
 \includegraphics[width=1.02\textwidth]{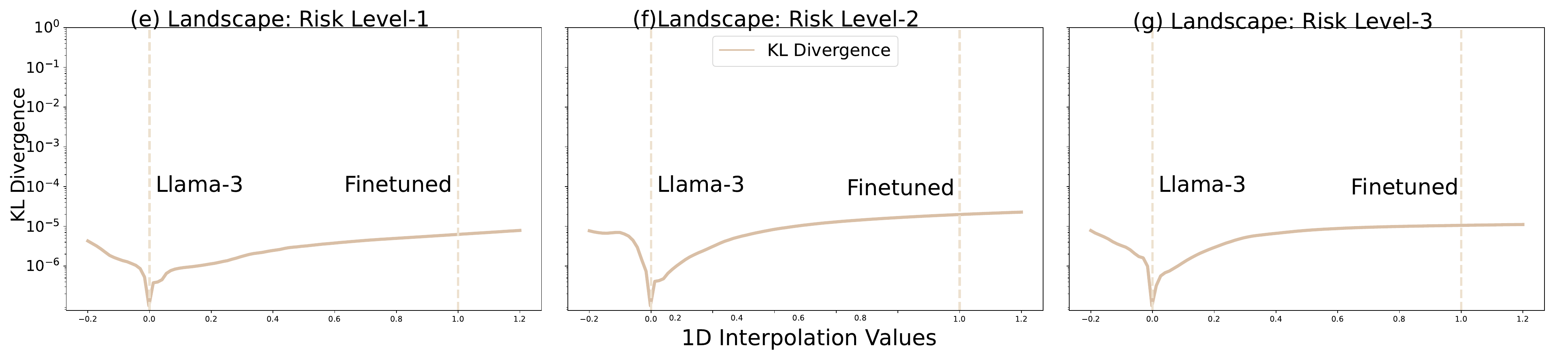}
\end{center}
 \caption{Analysis of safety directions at the 31st layer, 26th head for the original and fine-tuned models under different risk levels. (a-d) Activation density distributions. (e-g) KL divergence plots.}
 \label{Fig3}
\end{figure}

We examined the safety directions $\boldsymbol{\theta}_l^h$ in both the original Llama-3-Instruct 8B model and its fine-tuned variants subjected to different risk levels. Focusing on the most effective attention head (the 26th head in the 31st layer) for clarity, as depicted in Figure \ref{Fig3}, we observed distinct clustering of activations corresponding to safe (blue) and unsafe (orange) responses across both original and fine-tuned models. The black arrows in Figures \ref{Fig3}a-d illustrate that the shift from unsafe to safe activations maintains a high degree of similarity and consistency, regardless of the fine-tuning risk parameters applied. Additionally, our quantitative analysis using Kullback-Leibler (KL) divergence (Figure \ref{Fig3}e-g) revealed that the divergence between the original and fine-tuned models remains exceptionally low (below $10^{-5}$) across all tested risk levels. This minimal divergence indicates that the underlying safety-related activation patterns are largely preserved during fine-tuning. Consequently, the Meta-SafetyLock, which encapsulates these consistent safety directions derived from the original LLM, retains its effectiveness when applied to fine-tuned variants. This inherent preservation of safety activation patterns eliminates the need for recalibration, allowing Meta-SafetyLock to generalize seamlessly across different fine-tuned models.

\subsection{Distributing SafetyLock}
\label{sec:3.4}

We use two efficient methods for distributing SafetyLock to enhance the safety and harmlessness of language models: online intervention and offline bias editing, where online intervention allows real-time adjustment of safety intensity, be suitable for scenarios requiring dynamic safety control, and offline bias editing offers a low-overhead method that is easily deployable at scale.  

\textbf{Online Intervention.} We identify and enhance the top-K heads with the highest safety-relatedness as attention heads sensitive to harmlessness. For each of the select Top-K heads, we compute $\boldsymbol{\sigma}_{l}^h \in \mathbb{R}^D$, which represents the standard deviation of activations along each dimension of the safety direction $\boldsymbol{\theta}_{l}^h$. Specifically, we calculate:$\boldsymbol{\sigma}_{l}^h = \text{std}\left( \left\{ \mathbf{x}_{l}^h \odot \boldsymbol{\theta}_{l}^h \right\}_{i=1}^N \right)$. Where $\odot$ denotes element-wise multiplication, and $\text{std}$ computes the standard deviation across all $N$ samples for each dimension $d \in \{1, \dots, D\}$. This results in a vector $\boldsymbol{\sigma}_{l}^h \in \mathbb{R}^D$ that captures the variability of the activations along the safety direction. We modify the model's computation by adding a scaled version of the safety vector to the attention outputs for each select head:
\vspace{-0.1cm}
\begin{equation}
x_{l+1} = x_l + \sum_{h=1}^H Q_l^h \left( \operatorname{Att}_l^h(P_l^h x_l) + \alpha \boldsymbol{\sigma}_{l}^h \theta_l^h\right),
\end{equation}

where $\alpha$ controls safety intensity, the process is integrated into the autoregressive prediction for each subsequent token. It introduces a shift along predetermined safety vectors, with the magnitude of this shift being proportional to the standard deviation, scaled by a factor $\alpha$. 

\paragraph{Offline Bias Editing.} We can also modify the model's bias terms in an one-time manner:
\vspace{-0.1cm}
\begin{equation}
\text{Bias}_l = \text{Bias}_l + \alpha \sum_{h=1}^H Q_l^h \left(\sigma_l^h \theta_l^h\right).
\end{equation}
\vspace{-0.1cm}

\section{Experiments}
In this section, we present experiments to evaluate the effectiveness of the SafetyLock in enhancing model safety and inference efficiency, while maintaining model's general performance. We specifically address the following research questions:
\begin{itemize}[leftmargin=12pt]
  \item Can SafetyLock simultaneously improve the LLM's safety over all risk levels? (Section \ref{sec:4.1})
  \item What advantages does SafetyLock offer over post-training, inference methods? (Section \ref{sec:4.2},\ref{sec:4.3})
  \item How does SafetyLock reconcile the inherent trade-off between maintaining general capabilities and ensuring harmlessness in language models? (Section \ref{sec:4.4})
\end{itemize}

\subsection{Experimental Details}
\textbf{Threat Model Selections}. Following previous red teaming and safeguarding studies on aligned LLMs \citep{yuan2024refuse}, we consider a threat model where attackers can fine-tune aligned LLMs, typically through API access to closed-source models. The primary objective is jailbreaking these models and removing safety constraints \citep{wei2023jailbroken, carlini2023aligned} while SafetyLock aims to rebuild the safety guard. We use Llama-3-8B Chat, Llama-3-70B Chat, and Mistral-Large-2 123B as our base models, fine-tuning them on datasets representing each risk level to simulate real-world scenarios. Please refer to Appendix \ref{appendix:4} for detailed baseline experimental setups.

\textbf{Fine-tuning Datasets}. We conducted experiments on three risks: (1) explicitly harmful datasets, where attackers intentionally fine-tune models on malicious content \citep{ganguli2022redteaminglanguagemodels,qi2023visual}; (2) implicitly harmful datasets, which may appear benign but lead to compromised safety guardrails \citep{Qi2023FinetuningAL}; and (3) benign datasets, where even well-intentioned fine-tuning can inadvertently degrade model safety \citep{wang2023selfinstructaligninglanguagemodels}. For Risk-1, we use negative samples from the HH-RLHF preference dataset \citep{bai2022training}. We select 10, 100, 1000, and 10000 samples respectively and trained for 5 epochs with a learning rate of 2e-5. For Risk-2, we use 10 samples from \citet{Qi2023FinetuningAL} and train for 5 epochs with a learning rate of 2e-5. For Risk-3, we used the first 50,000 samples from the Alpaca dataset \citep{wang2023selfinstructaligninglanguagemodels} and trained for 5 epochs with a learning rate of 2e-5. 

\textbf{Safety Evaluation and Metrics}. Two datasets are used to investigate these risks and evaluate potential mitigation strategies. 
HEx-PHI \citep{Qi2023FinetuningAL} is based on 11 categories of prohibited use cases merged from Meta's Llama-3 acceptable use policy and OpenAI's usage policies. The dataset includes 30 examples per category, totalling 330 examples. This ensures a comprehensive safety evaluation aligned with industry-standard usage policies. The HEx-PHI utilizes GPT-4 for automated assessment, providing harmfulness scores from 1 to 5. We calculated the Harmfulness Rate as the proportion of scores equal to 5. AdvBench is released by \cite{zou2023universaltransferableadversarialattacks}, we adhere to the original paper's setup and calculate the ASR through string matching. 

\textbf{Baselines}. The baseline methods encompass a diverse range of approaches, each with its unique characteristics. Inference-time methods include ICD \citep{wei2024jailbreakguardalignedlanguage}, PPL \citep{alon2023detectinglanguagemodelattacks}, Paraphrase \citep{jain2023baselinedefensesadversarialattacks}, Retokenization \citep{jain2023baselinedefensesadversarialattacks}, Self-Reminder \citep{Xie2023DefendingCA}, and Self-Examination \citep{phute2024llm}, which operate without modifying the underlying model. Training-based methods, such as PPO, DPO, SFT with safety data mixing, and Model-Edited (DINM) \citep{wang2023easyedit,wang2024SafeEdit}, involve altering the model's parameters to enhance safety. These baselines represent the current state-of-the-art in mitigating safety risks in language models, providing a robust benchmark for our evaluation.

\subsection{Results over Different Risk Levels}
\label{sec:4.1}

\begin{figure}[h]
\centering
\includegraphics[width=\textwidth]{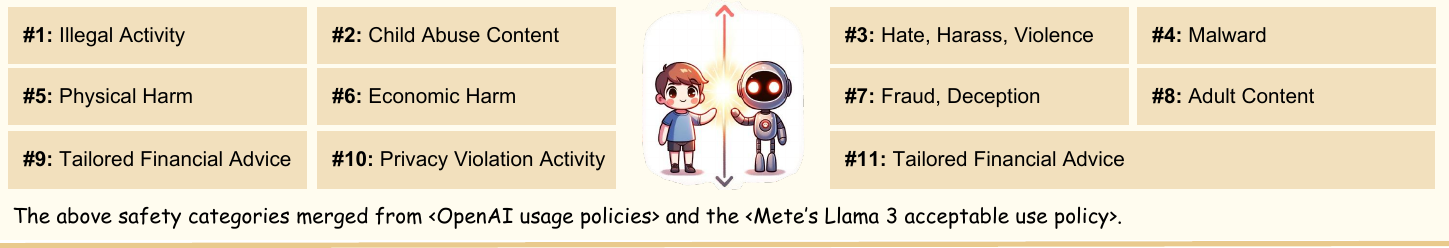}
\vspace{-0.3cm}
\includegraphics[width=\textwidth]{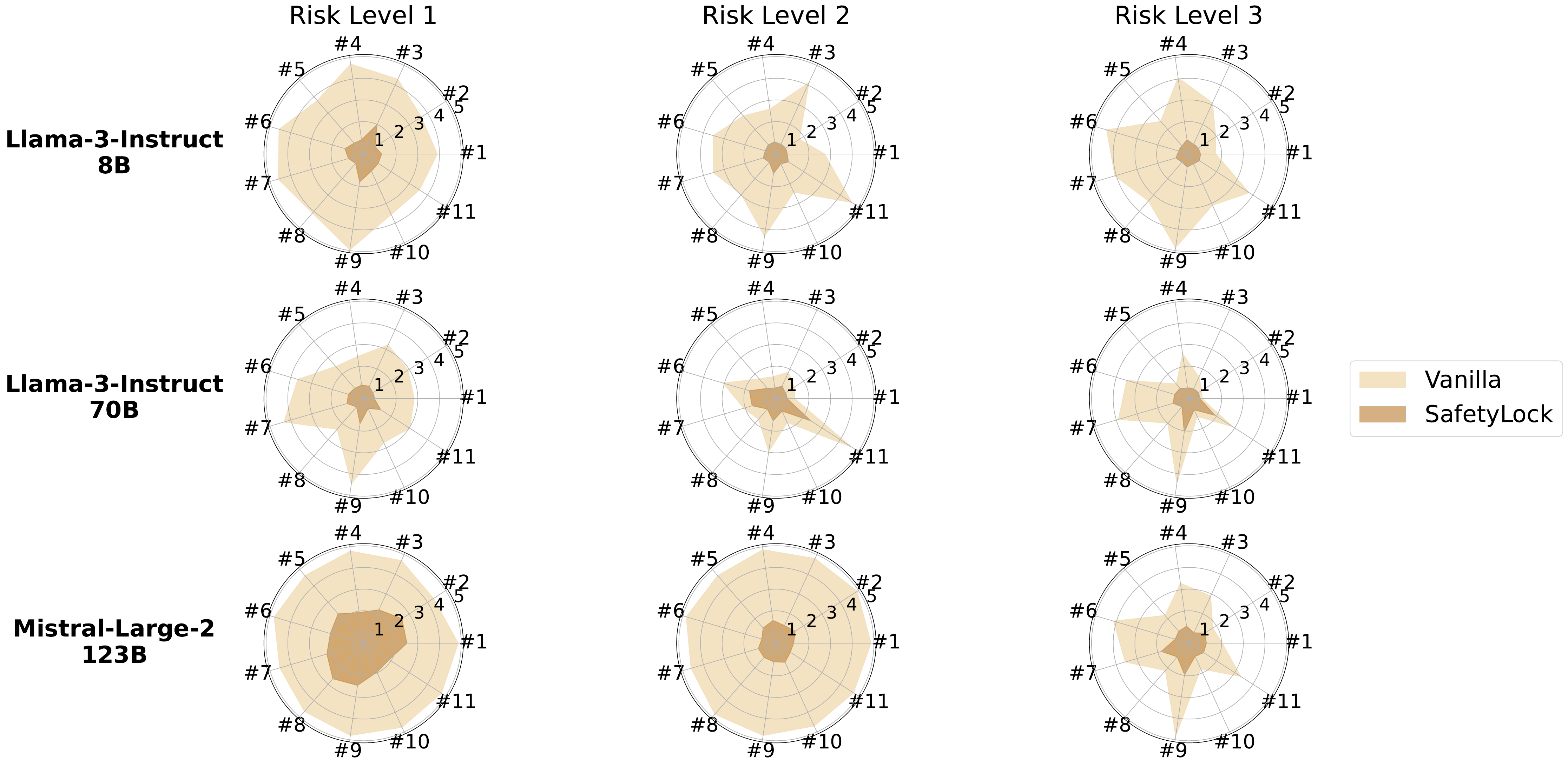}
\vspace{-0.3cm}
\caption{Safety performance comparison for 3 Risk Levels fine-tuned LLMs. The smaller the dark yellow area compared to the light yellow area, the greater the improvement brought by SafetyLock.}
\vspace{-0.1cm}
\label{fig:level1}
\end{figure}

\begin{table}[h!]
\centering
\caption{Comparison of Llama-3-8B-Instruct and Llama-3-70B-Instruct models for Risk 1, Risk 2, and Risk 3 scenarios. `Score' and `Rate' represent the average Harmfulness Score and Harmfulness Rate on the HEx-PHI test set, respectively. `ASR' denotes the Attack Success Rate on AdvBench.}
\vspace{0.3cm}
\scalebox{0.758}{
\begin{tabular}{l|c|ccc|ccc|ccc}
\toprule[1pt]
\multirow{2}{*}[-0.5ex]{Model} & \multirow{2}{*}[-0.5ex]{Method} & \multicolumn{3}{c|}{Risk 1: Explicitly harmful} & \multicolumn{3}{c|}{Risk 2: Identity Shifting} & \multicolumn{3}{c}{Risk 3: Benign} \\ [0.5ex] \cline{3-11} 
 &  & Score & Rate & ASR & Score & Rate & ASR & Score & Rate & ASR \\ [0.5ex]
\midrule[0.75pt]
\multirow{2}{*}[-0.5ex]{\textbf{\large \textit{\begin{tabular}[c]{@{}l@{}}Llama-3-8B-\\ Instruct\end{tabular}}}} & Vanilla & 4.13 & 70.01\% & 49.24\% & 3.19 & 53.33\% & 38.46\% & 3.23 & 54.24\% & 42.88\% \\ [0.5ex]
& \cellcolor[HTML]{F9EBE3}\textbf{SafetyLock} & \cellcolor[HTML]{F9EBE3}\textbf{1.36} & \cellcolor[HTML]{F9EBE3}\textbf{3.33\%} & \cellcolor[HTML]{F9EBE3}\textbf{0.19\%} & \cellcolor[HTML]{F9EBE3}\textbf{1.07} & \cellcolor[HTML]{F9EBE3}\textbf{1.21\%} & \cellcolor[HTML]{F9EBE3}\textbf{5.19\%} & \cellcolor[HTML]{F9EBE3}\textbf{1.04} & \cellcolor[HTML]{F9EBE3}\textbf{0.03\%} & \cellcolor[HTML]{F9EBE3}\textbf{0.19\%} \\ [0.5ex]
\midrule[0.75pt]
\multirow{2}{*}[-0.5ex]{\textbf{\large \textit{\begin{tabular}[c]{@{}l@{}}Llama-3-70B-\\ Instruct\end{tabular}}}} & Vanilla & 3.11 & 45.76\% & 44.81\% & 2.12 & 15.63\% & 9.42\% & 2.26 & 30.61\% & 20.77\% \\ [0.5ex]
& \cellcolor[HTML]{F9EBE3}\textbf{SafetyLock} & \cellcolor[HTML]{F9EBE3}\textbf{1.16} & \cellcolor[HTML]{F9EBE3}\textbf{3.64\%} & \cellcolor[HTML]{F9EBE3}\textbf{3.33\%} & \cellcolor[HTML]{F9EBE3}\textbf{1.30} & \cellcolor[HTML]{F9EBE3}\textbf{5.58\%} & \cellcolor[HTML]{F9EBE3}\textbf{1.67\%} & \cellcolor[HTML]{F9EBE3}\textbf{1.22} & \cellcolor[HTML]{F9EBE3}\textbf{5.15\%} & \cellcolor[HTML]{F9EBE3}\textbf{1.15\%} \\ [0.5ex]

\midrule[0.75pt]
\multirow{2}{*}[-0.5ex]{\textbf{\large \textit{\begin{tabular}[c]{@{}l@{}}Mistral-Large-2\\ 123B\end{tabular}}}} & Vanilla & 4.71 & 85.45\% & 80.77\% & 4.79 & 92.12\% & 82.50\% & 2.84 & 49.09\% & 19.23\% \\ [0.5ex]
& \cellcolor[HTML]{F9EBE3}\textbf{SafetyLock} & \cellcolor[HTML]{F9EBE3}\textbf{2.28} & \cellcolor[HTML]{F9EBE3}\textbf{1.52\%} & \cellcolor[HTML]{F9EBE3}\textbf{16.92\%} & \cellcolor[HTML]{F9EBE3}\textbf{1.38} & \cellcolor[HTML]{F9EBE3}\textbf{0\%} & \cellcolor[HTML]{F9EBE3}\textbf{10.00\%} & \cellcolor[HTML]{F9EBE3}\textbf{1.35} & \cellcolor[HTML]{F9EBE3}\textbf{5.15\%} & \cellcolor[HTML]{F9EBE3}\textbf{1.82\%} \\ [0.5ex]
\bottomrule[1pt]
\end{tabular}
}
\label{tab:llama-risk-comparison-merged}
\end{table}

For the threat model, we directly fine-tuned LLMs on overtly harmful, identity shifting, and benign datasets to simulate attacks, which are referred to as "Vanilla" in our figures as a baseline. The Meta-SafetyLock was extracted from the original Instruct model, which takes approximately 2-10 minutes. Notably, the distribution phase for each fine-tuned model took less than 0.01 seconds.

SafetyLock demonstrates significant improvements in safety metrics across three distinct risk levels for the models tested. Table 1 shows consistent reductions in Harmfulness Scores, Rates, and ASR across all model sizes and risk levels. 

\begin{wrapfigure}{t}{0.55\textwidth}
\begin{center}
\vspace{0.1cm}
\includegraphics[width=0.54\textwidth]{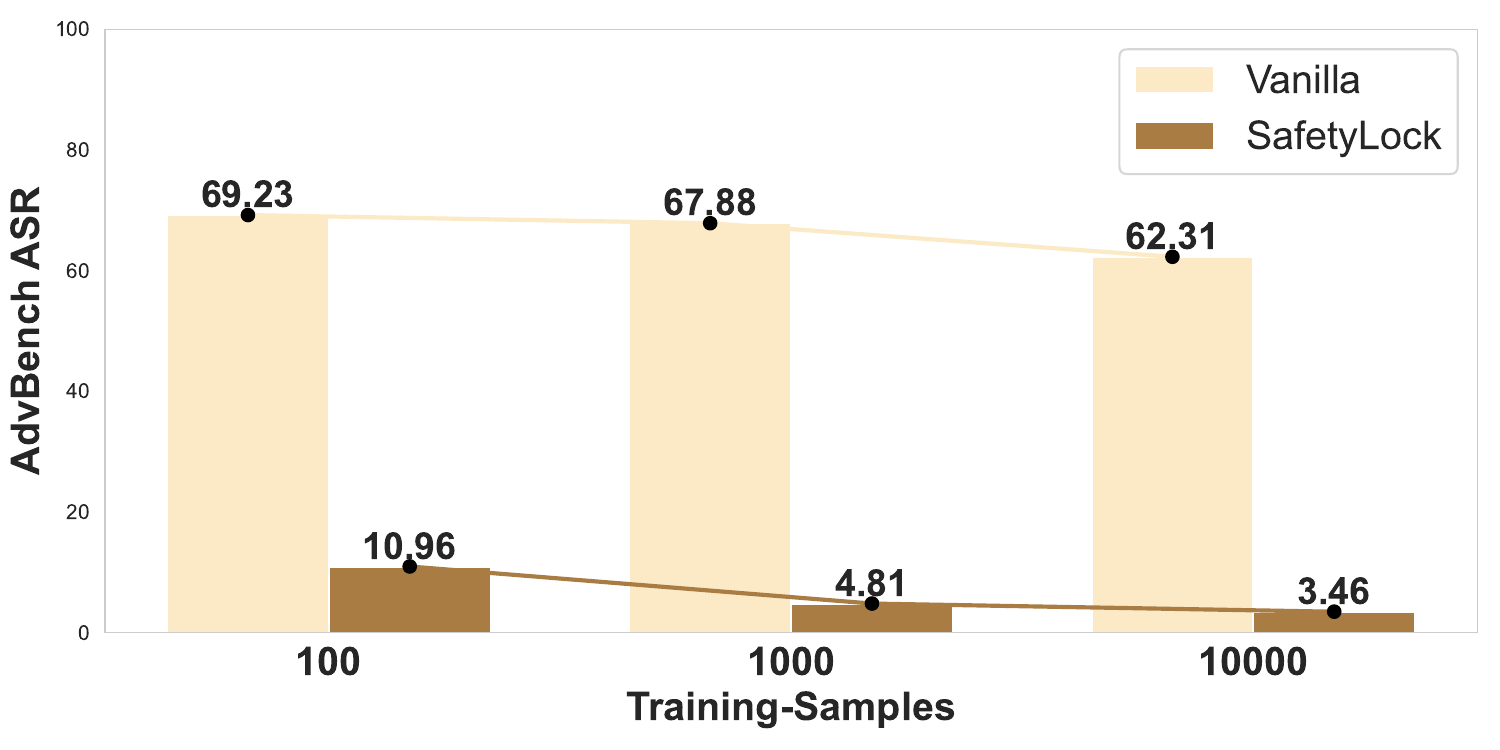}
\end{center}
\vspace{0.1cm}
\caption{Impact of increasing harmful training samples on model safety with and without SafetyLock.}
\label{fig:scalability}
\end{wrapfigure}

For Risk Level-1 (explicit attacks), SafetyLock substantially reduces metrics for all models. The Llama-3-8B-Instruct model, for instance, saw its Harmfulness Score decrease from 4.13 to 1.36, Rate from 70.01$\%$ to 3.33$\%$, and ASR from 49.24$\%$ to 0.19$\%$ . Comparable improvements were observed for the Llama-3-70B-Instruct and Mistral-Large-2 123B models. Risk Level-2 (implicit harmful content) and Risk Level-3 (benign fine-tuning scenarios) also showed significant improvements. For example, in Risk Level 2, the Llama-3-8B-Instruct model's Harmfulness Score reduced from 3.19 to 1.07, while in Risk Level 3, it decreased from 3.23 to 1.04. Similar improvements were observed across all model sizes, demonstrating SafetyLock's ability to maintain ethical guardrails during routine model customization processes. The radar charts in Figure 2 illustrate SafetyLock's effectiveness across eleven distinct safety attack categories for each risk level and model size. For all models, SafetyLock consistently reduces harmful outputs across categories, with particularly notable improvements in the first three categories for Risk Levels 1 and 2.

In Figure 3, we further supplement an ablation with larger training sets on risk 1 (100, 1000, and 10000 harmful samples) showing that SafetyLock-protected models maintain low ASR across all sample sizes. Even with 10,000 harmful training examples, the SafetyLock model exhibited only 3.46$\%$ ASR, compared to 62.31$\%$for the unprotected model. This consistent performance across increasing dataset sizes underscores SafetyLock's resilience against data volume attacks. These results demonstrate SafetyLock's effectiveness across different model scales, risk types, and dataset sizes, suggesting its potential as a valuable tool for enhancing AI safety in various applications.

\subsection{Comparative Analysis of Baseline Methods} 
\label{sec:4.2}

\begin{figure}[t]
\begin{center}
	\includegraphics[width=\textwidth]{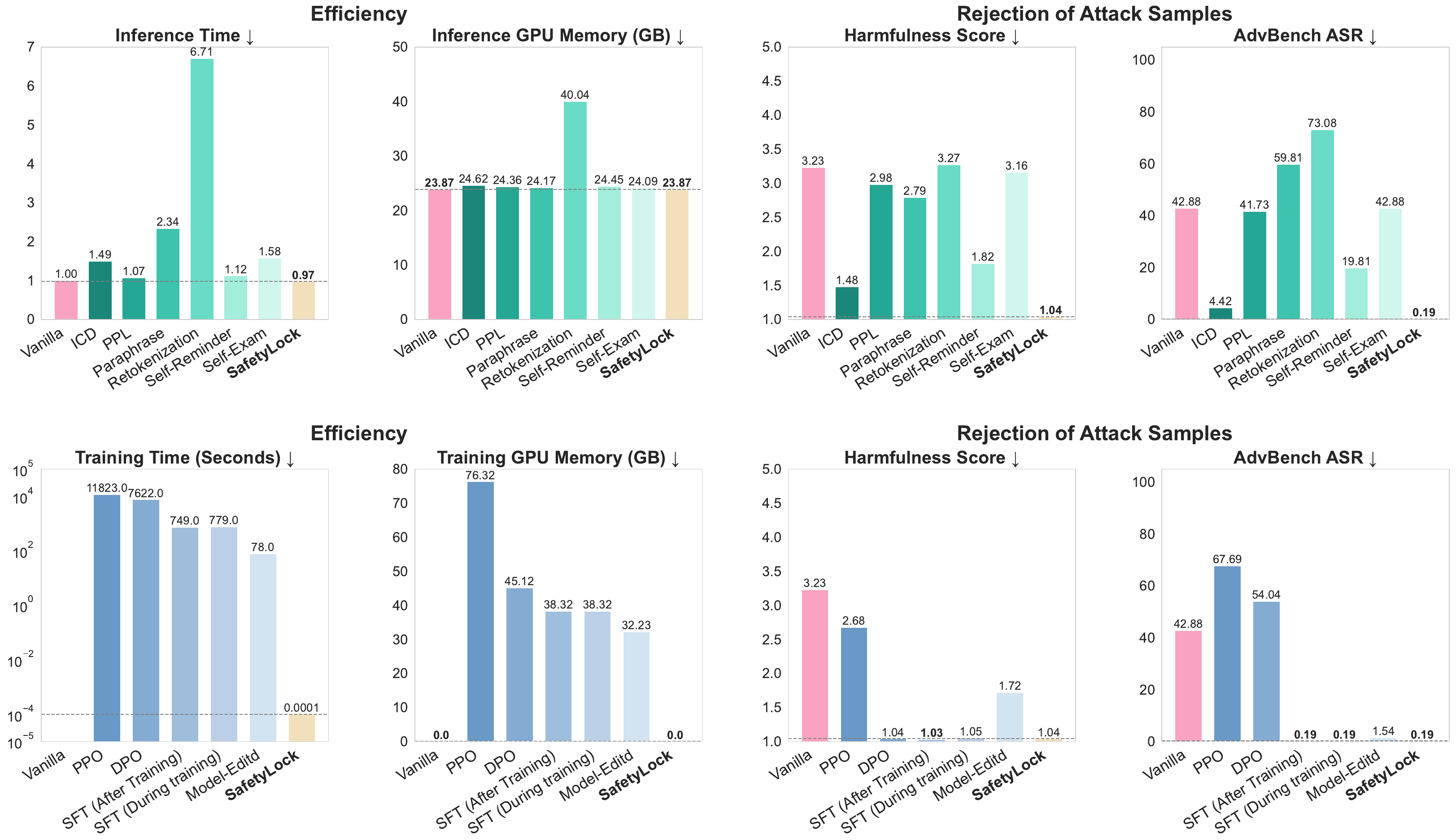}
\end{center}
 \caption{Comparison of Methods for Mitigating Safety Risks in Fine-tuned Language Models (Llama-3-Instruct 8B). \textbf{Upper row: Compared with inference-time methods; Lower row: Compared with training-time methods}, Each row represents efficiency metrics(training time and GPU memory), and rejection of attack samples (Harmfulness Score and AdvBench ASR).}
 \vspace{-0.2cm}
 \label{fig:comparison}
\end{figure}
To comprehensively evaluate SafetyLock's efficacy, we conducted a comparative analysis against established baseline methods, categorized into training-based and inference-time approaches, as illustrated in Figure \ref{fig:comparison}. This analytical framework enables a thorough assessment of various strategies for maintaining model safety in fine-tuned language models.

As demonstrated in Figure \ref{fig:comparison}, in terms of efficiency, SafetyLock exhibits a remarkable computational economy. Its inference time of 0.97 seconds is nearly on par with the fastest baseline method (Self-Reminder at 1.12 seconds), while its training time of 0.01 seconds and additional GPU memory usage of 0.0 GB are orders of magnitude lower than all training-based methods. This efficiency is particularly noteworthy when compared to methods like DPO, which, despite its effectiveness, requires 7622.0 seconds of training time and 45.12 GB of GPU memory. Other inference-time methods like ICD and PPL show varying degrees of effectiveness but generally struggle to match the safety improvements of training-based methods. SFT with safety data mixing post-fine-tuning offers a more balanced approach, achieving a Harmfulness Score of 1.03 with reduce resource requirements of 779 seconds and 38.32 GB GPU memory. 
Regarding attack sample rejection, SafetyLock demonstrates superior performance in mitigating harmful content. It achieves a Harmfulness Score of 1.04, equivalent to that achieved by models undergoing safety realignment via DPO, indicating its exceptional ability to reduce the generation of harmful content. Furthermore, SafetyLock's AdvBench ASR of 0.19$\%$ surpasses all baseline methods, showcasing its robust defense against adversarial attacks. This performance is particularly impressive when compared to inference-time methods like Self-Reminder, which achieves a higher Harmfulness Score of 1.82 and an AdvBench ASR of 19.81\%.

We further assess the models' performance on benign inputs to ensure safety enhancements did not compromise normal text generation by selecting 500 test samples from the Alpaca dataset. The results reveal that SafetyLock preserves a 98.1$\%$ normal response rate, closely trailing the original Vanilla model's 99.4$\%$. Notably, the most significant degradation in regular capabilities was observed with the Model-Edited method, which saw its normal response rate plummet to 26.8$\%$. Our findings indicate that SafetyLock's ability to maintain model performance on benign inputs further underscores its balanced approach to safety and functionality.

In conclusion, \textbf{SafetyLock distinguishes itself by achieving an exceptional balance between efficiency and robust defense against harmful content, without compromising the model's ability to generate plausible responses.} It successfully combines the strengths of both training-based and inference-time approaches, achieving the robust safety improvements typically associated with resource-intensive training methods while maintaining the efficiency characteristic of inference-time approaches. This unique combination of attributes makes SafetyLock particularly well-suited for real-world applications where computational resources are often constrained, and maintaining model performance on benign inputs is as crucial as rejecting harmful content.

\subsection{SafetyLock's Performance Against Combined Attacks}
\label{sec:4.3}

\begin{table}[t]
\centering
\caption{Comparison of SafetyLock and other inference-time defence methods against four prominent prompt-based attacks on fine-tuned Llama-3-8B Instruct.}
\vspace{0.3cm}
{
\begin{tabular}{l|c|c|c|c}
\toprule
Model & AutoDAN ASR & DeepInception ASR & GCG ASR & PAIR ASR \\
\midrule
Vanilla & 84.0 & 98.0 & 74.0 & 70.0 \\
ICD & 46.0 & 98.0 & 22.0 & 50.0 \\
PPL & 84.0 & 98.0 & 0.0 & 70.0 \\
Paraphrase & 32.0 & 96.0 & 58.0 & 74.0 \\
Relexicalization & 82.0 & 98.0 & 94.0 & 64.0 \\
Self-Reminder & 66.0 & 98.0 & 32.0 & 56.0 \\
Star Exam & 84.0 & 98.0 & 74.0 & 70.0 \\
\textbf{SafetyLock} & \cellcolor[HTML]{F9EBE3}\textbf{4.0} & \cellcolor[HTML]{F9EBE3}\textbf{2.0} & \cellcolor[HTML]{F9EBE3}\textbf{10.0} & \cellcolor[HTML]{F9EBE3}\textbf{14.0} \\
\bottomrule
\end{tabular}
}
\label{tab:asr-comparison}
\end{table}

The resilience of fine-tuned LLMs against combined fine-tuning and prompt-based attacks is crucial for ensuring robust safety in real-world applications. To further assess robustness, we introduced a combined attack scenario: fine-tuning model attacks followed by prompt-based attacks. We evaluated four commonly use prompt attack methods: AutoDAN \citep{liu2024autodan}, DeepInception \citep{li2024deepinceptionhypnotizelargelanguage}, GCG \citep{zou2023universaltransferableadversarialattacks}, and PAIR \citep{chao2024jailbreakingblackboxlarge}, comparing their performance against several inference-time defense techniques, as illustrated in Table \ref{tab:asr-comparison}. 

SafetyLock demonstrates exceptional effectiveness across all tested attack methods. For AutoDAN attacks, SafetyLock reduces the ASR to a mere 4.0$\%$, significantly outperforming other methods such as ICD (46.0$\%$) and Self-Exam (66.0$\%$). Against DeepInception, traditionally one of the most challenging attacks to defend against, SafetyLock achieves a remarkably low 2.0$\%$ ASR, while all other methods fail to provide any meaningful defense (98.0$\%$ ASR across the board). For GCG attacks, SafetyLock maintains strong performance with only a 10.0$\%$ ASR, second only to PPL's 0.0$\%$ but considerably better than most other methods, including Vanilla (74.0$\%$) and Retokenization (94.0$\%$). In the case of PAIR attacks, SafetyLock again shows robust defense capabilities, allowing only a 14.0$\%$ ASR, outperforming all other tested methods.

\textbf{These results underscore SafetyLock's versatility and effectiveness in mitigating prompt-based attacks across various attack types.} Its consistent performance demonstrates a comprehensive approach to model safety, addressing the complex challenges posed by diverse attack scenarios in language model deployment. The ability to maintain such low ASR across different attack methods suggests that SafetyLock provides a more generalizable and robust defense mechanism.

\subsection{Generalization Capabilities of SafetyLock}
\label{sec:4.4}
\begin{figure}[h]
\begin{center}
	\includegraphics[width=\textwidth]{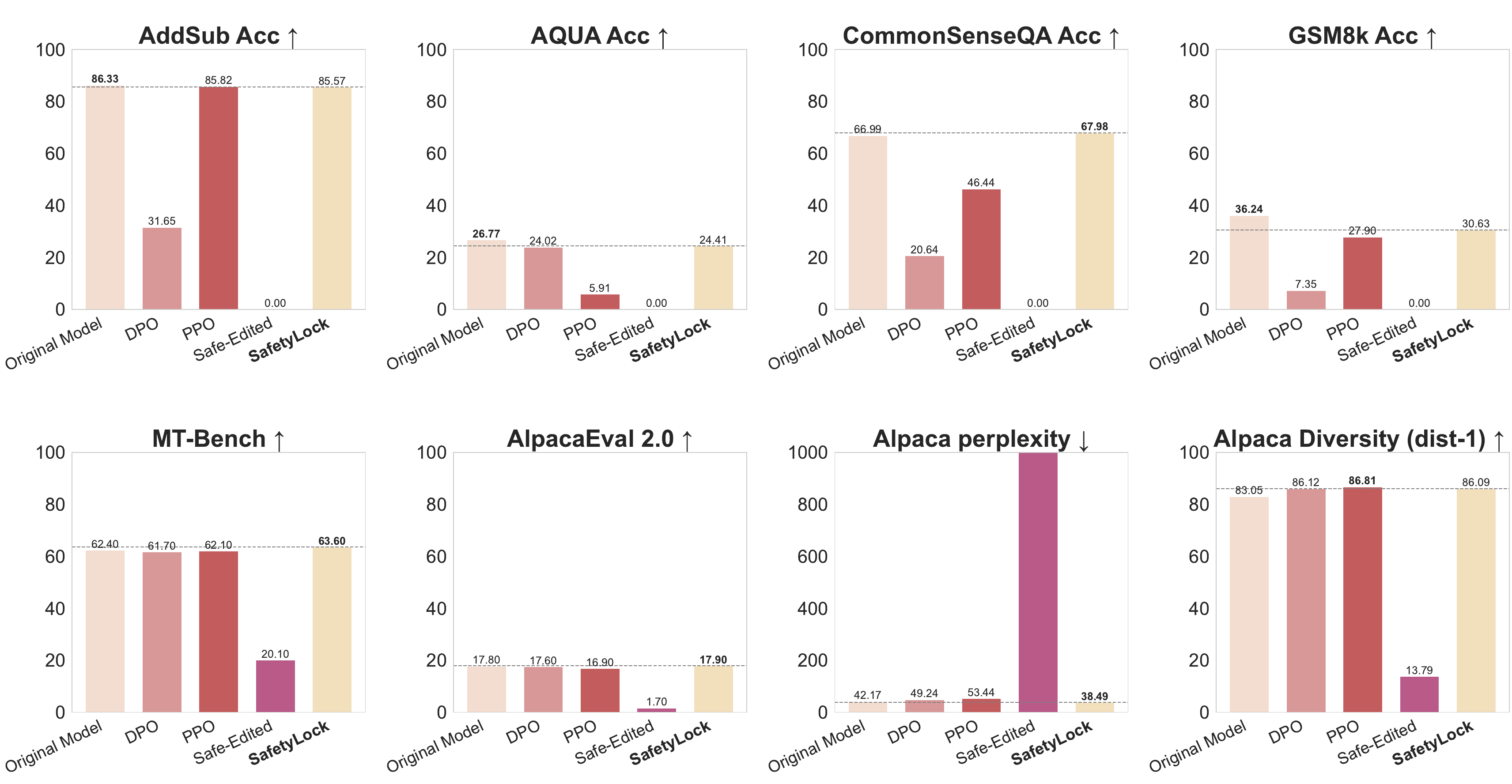}
\end{center}
 \vspace{-0.5cm}
 \caption{Performance comparison of various methods on downstream tasks.}
 \vspace{-0.2cm}
 \label{fig:pc}
\end{figure}

We assess a wide range of language understanding and generation capabilities to provide a comprehensive view of model performance based on various downstream tasks. Our experiments include a diverse set of benchmarks \citep{MohammadJavadHosseini2014LearningTS,AlonTalmor2018CommonsenseQAAQ,PatelArkil2021AreNM,cobbe2021training,suzgun2022challenging,SubhroRoy2016SolvingGA,wei2022chain,kojima2022large,weng2024mastering,zheng2023judging,dubois2023alpacafarm}: AddSub, AQUA, CommonSenseQA, GSM8k, MT-Bench, Alpaca, and AlpacaEval 2.0 .

As illustrated in Figure \ref{fig:pc}, SafetyLock demonstrates a remarkable ability to maintain model performance across all tasks while ensuring safety. Unlike previous knowledge editing methods, which often led to significant performance degradation or incoherent outputs, SafetyLock preserves the model's foundational capabilities. For instance, on the AddSub task, SafetyLock maintains a performance of 85.57\%, closely matching the original model's 86.33\%, while other methods like Model-Edited show complete performance collapse. This trend is consistent across other tasks, with SafetyLock consistently performing on par with or slightly below the original model, in stark contrast to the severe degradation seen with other safety-aligned methods. These results highlight SafetyLock's unique ability to enhance model safety without compromising its core functionalities, addressing a critical challenge in the deployment of safe and effective language models.

\section{Conclusion}

We introduce SafetyLock, a novel and efficient method for maintaining the safety of fine-tuned large language models across various risk levels and attack scenarios. Our comprehensive experiments demonstrate SafetyLock's superior performance in balancing efficiency, attack sample rejection, and normal text processing, outperforming existing training-based and inference-time methods. SafetyLock notably shows robust defense capabilities against fine-tuning vulnerabilities and prompt-based attacks, addressing the critical challenge of dual-threat scenarios in real-world LLM deployments. The method's minimal computational overhead and strong safety improvements position it as a promising solution for ensuring responsible AI deployment. Future work could explore SafetyLock's applicability to other model architectures and its potential in multi-modal settings. Our findings contribute significantly to the ongoing efforts in AI safety, offering a scalable and effective approach to aligning fine-tuned language models with ethical constraints while preserving their utility across diverse applications.

\newpage
\section*{Reproducibility Statement}

We have taken several steps to ensure the reproducibility of our results. The implementation details, datasets, and models used in our experiments are described in the corresponding sections of this paper, particularly in Sections \ref{sec:3.2}, \ref{sec:3.4}, and \ref{sec:4.1}. We also provide the experimental settings and evaluation metrics in Sections \ref{sec:3.3} and \ref{sec:4.2}. Furthermore, all hyperparameters, training code, and baselines are detailed throughout the relevant sections, ensuring that researchers can replicate our work using publicly available datasets and models.

\bibliography{iclr2024_conference}
\bibliographystyle{iclr2024_conference}

\appendix
\section{Appendix}

\subsection{Limitations}

While SafetyLock demonstrates promising results in maintaining the safety of fine-tuned language models, it is important to acknowledge several limitations. Primarily, SafetyLock requires access to both model weights and intermediate activations for implementation, which may limit its applicability in scenarios where such access is restricted or unavailable. Additionally, the method employs a symmetric locking mechanism; consequently, if an unauthorized party gains access to the model weights or activation values, they could potentially reverse-engineer the process to unlock and bypass SafetyLock's protections. Lastly, while SafetyLock shows strong performance against current attack methods, its long-term robustness against evolving adversarial techniques remains to be studied. These limitations present opportunities for future work to enhance and expand the capabilities of SafetyLock, ensuring its continued effectiveness in maintaining AI safety.

\subsection{Consistency of Harmlessness Directions in Fine-tuned Models}
\label{appendix:harmlessness_consistency}

To validate SafetyLock's effectiveness, we conducted a comprehensive analysis of the original Llama-3-Instruct 8B model and its fine-tuned versions under various risk levels. Our experimental setup was as follows:

We first extracted activation values from the 31st layer, 26th head of the Llama-3-8B Instruct model, which we identified as the most sensitive to harmlessness through linear regression, achieving the highest binary classification accuracy. We then performed forward computation on a safety dataset, saving the activation values of the last token for both safe and unsafe samples. Using 2D PCA for dimensionality reduction, we visualized the shift in activation values between safe and unsafe samples by connecting their center points with arrows, illustrating both the direction and magnitude of the shift.

Remarkably, we observed high similarity in these shifts across different risk levels (i.e., fine-tuning on data from different domains). To quantitatively assess the similarity between the safety directions found in the original model and those in the fine-tuned models, we employed KL divergence:

\begin{equation}
D_{KL}(P || Q) = \sum_{i} P(i) \log \left(\frac{P(i)}{Q(i)}\right)
\end{equation}

where $P$ and $Q$ represent the distributions of safety directions in the original and fine-tuned models, respectively.

To further illustrate the change in similarity during the fine-tuning process, we employed one-dimensional linear interpolation of weights \citep{peng2024navigatingsafetylandscapemeasuring}. This method allows us to smoothly transition from the original model weights to the fine-tuned model weights, providing insight into how the safety directions evolve during the fine-tuning process. The interpolation is defined as:

\begin{equation}
\theta_{\alpha} = \theta + \alpha(\theta' - \theta)
\end{equation}

where $\theta$ represents the weights of the original Llama-3 model, $\theta'$ the weights of the fine-tuned model, and $\alpha \in [-0.2, 1.2]$ is the interpolation parameter. We extend $\alpha$ slightly beyond the [0, 1] range to observe trends slightly before and after the actual interpolation points.

The interpolation process is implemented as follows:

\begin{enumerate}
\item We first extract the state dictionaries of both the base model ($\theta$) and the fine-tuned model ($\theta'$).
\item For each layer, we compute the difference vector: $d_1 = \theta' - \theta$.
\item We then create new weights for each $\alpha$ value: $\theta_{\alpha} = \theta + \alpha d_1$.
\item These new weights are used to reconstruct a new state dictionary, maintaining the original structure and naming conventions of the model.
\end{enumerate}

We use these interpolated models to compute the KL divergence between the safety directions of the original model and the interpolated models at each step. This results in a smooth curve showing how the similarity of safety directions changes as the model transitions from its original state to the fine-tuned state.

\section{Mathematical Explanation of SafetyLock's Effectiveness in Suppressing Harmful Outputs}
\label{sec:theoretical_explanation}

In this section, we provide a mathematical justification for why SafetyLock can extract transferable safety directions from the original language model and effectively apply them to fine-tuned models to suppress harmful outputs. Our explanation is grounded in the properties of Transformer-based language models and the nature of fine-tuning on limited datasets.

\subsection{Activation Space and Safety Directions}

Let us denote the activations of the original (pre-fine-tuned) language model at layer $l$ and head $h$ as $\mathbf{x}_{l,h} \in \mathbb{R}^{D}$, where $D$ is the dimensionality of the head's output. During inference, these activations encode information about the generated tokens.

We define two sets of activations corresponding to safe and unsafe responses:

\begin{align}
    \mathcal{X}_{\text{safe}} &= \left\{ \mathbf{x}_{l,h}^{\text{safe}, i} \right\}_{i=1}^{N_{\text{safe}}}, \\
    \mathcal{X}_{\text{unsafe}} &= \left\{ \mathbf{x}_{l,h}^{\text{unsafe}, i} \right\}_{i=1}^{N_{\text{unsafe}}},
\end{align}

where $N_{\text{safe}}$ and $N_{\text{unsafe}}$ are the numbers of safe and unsafe samples, respectively.

We compute the \emph{safety direction} $\boldsymbol{\theta}_{l,h} \in \mathbb{R}^{D}$ as the mean difference between the activations for safe and unsafe responses:

\begin{equation}
    \boldsymbol{\theta}_{l,h} = \frac{1}{N_{\text{safe}}} \sum_{i=1}^{N_{\text{safe}}} \mathbf{x}_{l,h}^{\text{safe}, i} - \frac{1}{N_{\text{unsafe}}} \sum_{i=1}^{N_{\text{unsafe}}} \mathbf{x}_{l,h}^{\text{unsafe}, i}.
    \label{eq:safety_direction}
\end{equation}

This vector represents the average shift in activation space needed to move from an unsafe response towards a safe one.

\subsection{Preservation of Safety Directions During Fine-Tuning}

Fine-tuning a language model on a new dataset modifies its parameters to adapt to specific tasks or domains. However, when the fine-tuning dataset is limited in size or scope, the changes to the model's internal representations are often localized and do not significantly alter the global structure of the activation space \citep{golechha2024challengesmechanisticallyinterpretingmodel, NEURIPS2022_4df3510a}.

Let $\mathbf{\tilde{x}}_{l,h}$ denote the activations of the fine-tuned model at layer $l$ and head $h$. Empirically, we observe that there exists a strong linear relationship between the activations of the original and fine-tuned models:

\begin{equation}
    \mathbf{\tilde{x}}_{l,h} \approx \mathbf{x}_{l,h} + \Delta \mathbf{x}_{l,h},
    \label{eq:activation_relationship}
\end{equation}

where $\Delta \mathbf{x}_{l,h}$ represents the change in activations due to fine-tuning, which is relatively small in magnitude compared to $\mathbf{x}_{l,h}$ for many dimensions.

Moreover, the safety direction $\boldsymbol{\theta}_{l,h}$ computed from the original model remains relevant in the fine-tuned model because the relative differences between safe and unsafe activations are preserved:

\begin{equation}
    \boldsymbol{\tilde{\theta}}_{l,h} = \left( \mathbf{\tilde{x}}_{l,h}^{\text{safe}} - \mathbf{\tilde{x}}_{l,h}^{\text{unsafe}} \right) \approx \left( \mathbf{x}_{l,h}^{\text{safe}} - \mathbf{x}_{l,h}^{\text{unsafe}} \right) = \boldsymbol{\theta}_{l,h}.
    \label{eq:safety_direction_preservation}
\end{equation}

This approximation holds under the assumption that fine-tuning does not disproportionately affect the dimensions critical for encoding safety-related information.

\subsection{Effectiveness of Activation Intervention}

During inference with the fine-tuned model, we intervene by adjusting the activations along the safety direction:

\begin{equation}
    \mathbf{\tilde{x}}_{l,h}^{\text{intervened}} = \mathbf{\tilde{x}}_{l,h} + \alpha \left( \boldsymbol{\sigma}_{l,h} \odot \boldsymbol{\theta}_{l,h} \right),
    \label{eq:activation_intervention}
\end{equation}

where:

\begin{itemize}
    \item $\alpha \in \mathbb{R}$ is the scaling factor controlling the intensity of the intervention.
    \item $\boldsymbol{\sigma}_{l,h} \in \mathbb{R}^{D}$ is the standard deviation vector of activations along each dimension, capturing the typical variability.
    \item $\odot$ denotes element-wise multiplication.
\end{itemize}

This adjustment effectively shifts the activations towards regions in the activation space associated with safe responses. Since the safety direction $\boldsymbol{\theta}_{l,h}$ is approximately preserved in the fine-tuned model, this intervention remains effective.

\subsection{Impact on Output Probabilities}

The language model generates the next token based on a probability distribution computed from the final activations. Adjusting the activations as in Equation \eqref{eq:activation_intervention} influences the logits $\mathbf{z} \in \mathbb{R}^{V}$ (where $V$ is the vocabulary size) before the softmax function:

\begin{equation}
    \mathbf{z}^{\text{intervened}} = \mathbf{z} + W_{\text{head}} \left( \alpha \left( \boldsymbol{\sigma}_{l,h} \odot \boldsymbol{\theta}_{l,h} \right) \right),
    \label{eq:logit_adjustment}
\end{equation}

where $W_{\text{head}} \in \mathbb{R}^{V \times D}$ is the weight matrix projecting activations to logits.

The adjustment $\Delta \mathbf{z} = W_{\text{head}} \left( \alpha \left( \boldsymbol{\sigma}_{l,h} \odot \boldsymbol{\theta}_{l,h} \right) \right)$ biases the logits towards tokens that are more likely in safe responses and away from those prevalent in unsafe responses.

\subsection{Suppressing Harmful Outputs}

The probability of generating a harmful token $t_{\text{harm}}$ is given by:

\begin{equation}
    P(t_{\text{harm}}) = \frac{\exp\left( z_{t_{\text{harm}}}^{\text{intervened}} \right)}{\sum_{i=1}^{V} \exp\left( z_{i}^{\text{intervened}} \right)}.
    \label{eq:harmful_token_probability}
\end{equation}

By decreasing $z_{t_{\text{harm}}}^{\text{intervened}}$ relative to other logits, we reduce $P(t_{\text{harm}})$. Since the intervention shifts the activations towards safe regions, the logits for harmful tokens are decreased, and the model is less likely to generate harmful outputs.

\subsection{Transferability Across Models}

The key to SafetyLock's transferability lies in the similarity of safety directions between the original and fine-tuned models. Since the fine-tuning process does not significantly alter the relative positions of safe and unsafe activations in the activation space (as per Equation \eqref{eq:safety_direction_preservation}), the safety directions computed from the original model remain effective when applied to the fine-tuned model.

This property is supported by empirical observations of low Kullback–Leibler (KL) divergence between the activation distributions of the original and fine-tuned models (see Figure~\ref{Fig3} in Section~\ref{sec:3.3}). The minimal divergence indicates that the overall structure of the activation space, especially along dimensions relevant to safety, is preserved during fine-tuning.

\subsection{Conclusion}

Mathematically, SafetyLock leverages the preserved safety directions in the activation space to adjust the model's internal computations towards generating safe outputs. By intervening along these directions, we effectively suppress harmful responses without requiring retraining or fine-tuning of the model. The minimal changes to the activation distributions during fine-tuning ensure that the safety directions remain applicable, allowing for efficient and transferable safety interventions across different models and fine-tuning scenarios.

This theoretical explanation provides a foundation for understanding the effectiveness of SafetyLock in suppressing harmful outputs while maintaining the model's overall performance on benign tasks.

\section{The Risks of Fine-tuning LLMs and Experimental Setup}
\label{appendix:4}

HEx-PHI \citep{Qi2023FinetuningAL} is based on 11 categories of prohibited use cases merged from Meta's Llama-3 acceptable use policy and OpenAI's usage policies: (1) Illegal Activity, (2) Child Abuse Content, (3) Hate, Harass, Violence, (4) Malware, (5) Physical Harm, (6) Economic Harm, (7) Fraud, Deception, (8) Adult Content, (9) Political Campaigning, (10) Privacy Violation Activity, and (11) Tailored Financial Advice. The dataset includes 30 examples per category, totaling 330 examples. This ensures a comprehensive safety evaluation aligned with industry-standard usage policies.

For Risk-1, we use negative samples from the HH-RLHF preference dataset. We select 10, 100, 1000, and 10000 samples respectively and trained for 5 epochs with a learning rate of 2e-5. For Risk-2, we use 10 samples from \citet{Qi2023FinetuningAL} and trained for 5 epochs with a learning rate of 2e-5. For Risk-3, we use the first 50,000 samples from the Alpaca dataset \citep{wang2023selfinstructaligninglanguagemodels} and trained for 5 epochs with a learning rate of 2e-5 \footnote{We use the official fine-tuning code \url{https://github.com/meta-llama/llama-recipes}}. 

Recognizing the potential of existing approaches to address safety issues in fine-tuned language models, we conducted comparative analyses across two categories as the same time: training-based and inference-time methods. For training-based approaches, we evaluated PPO, DPO, SFT (with safety data mixed during fine-tuning), SFT (with safety data mixed post-fine-tuning), and model-editing. Inference-time methods included ICD, PPL, Paraphrase, Retokenization, Safe-Reminder, and Self-Exam. These methods were assess based on efficiency, attack sample rejection rate, and normal text rejection rate, providing a comprehensive evaluation of their effectiveness in maintaining model safety while preserving functionality. This multi-faceted approach allows us to rigorously examine the trade-offs between safety and performance.

Specifically, to ensure reproducibility, we followed past experimental settings and use 2000 safety data points from \citet{bianchi2024safetytuned} for SFT experiments. We considered two experimental settings for SFT. The first is After Training, which simulates the scenario where safety disappears after fine-tuning the language model and needs to be restored. This applies to all fine-tuned language models. The second is During Training, which simulates starting from the original model and requiring the mixing of additional safety data during training to prevent safety disappearance. However, the limitation of this method is that it still requires retraining for already fine-tuned language models. For PPO, we also use 2000 samples from \citet{bianchi2024safetytuned}, and we use LlamaGuard-7b \citep{bhatt2023purplellamacybersecevalsecure} as the Reward model. For DPO, based on the 2000 samples, we use samples generated by the fine-tuned language model (almost all of which are harmful) as negative samples for training. For the Model-Edited method, we use the most common Detoxifying with Intraoperative Neural Monitoring (DINM) method and followed the original setup using SafeEdit data\footnote{\url{https://huggingface.co/datasets/zjunlp/SafeEdit}} for editing.

\section{Analysis of SafetyLock's Intervention}
\label{appendix:D}

\begin{figure}[h]
\begin{center}
	\includegraphics[width=\textwidth]{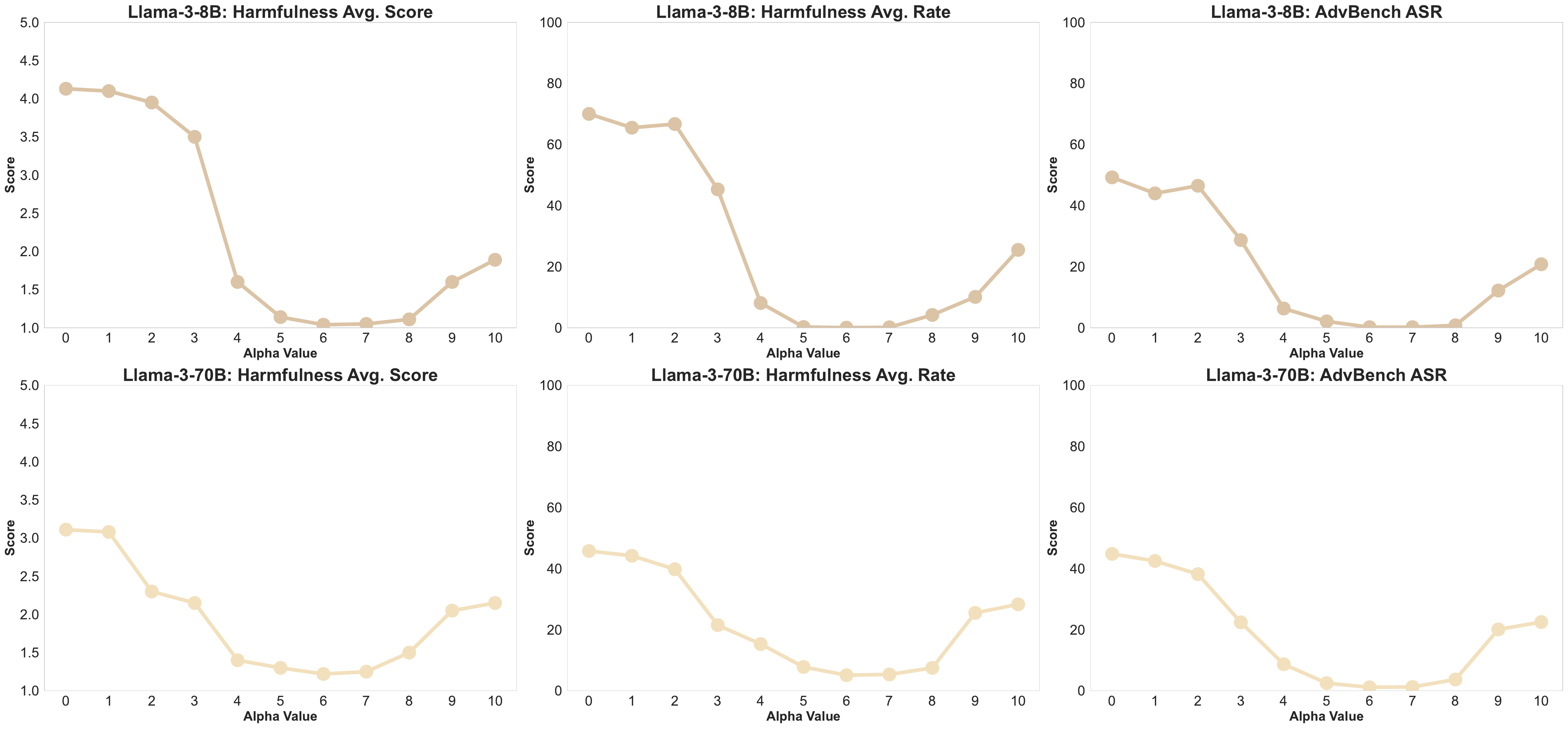}
\end{center}
\caption{Impact of SafetyLock's intervention distance ($\alpha$) on model safety metrics for Llama-3-8B and Llama-3-70B models. The graphs show Harmfulness Average Score, Harmfulness Average Rate, and AdvBench ASR across different $\alpha$ values. Note that for these experiments, the intervention degree K is set to 24, indicating the number of attention heads influenced by SafetyLock.}
\label{fig:alpha_impact}
\end{figure}

\textbf{Distance $\alpha$.} Our experimental results, as illustrated in Figure \ref{fig:alpha_impact}, demonstrate the significant influence of SafetyLock's intervention distance ($\alpha$) on model safety across different model sizes. For both Llama-3-8B and Llama-3-70B, we observe a clear U-shaped trend in harmfulness metrics as $\alpha$ increases. Initially, as $\alpha$ rises from 0 to 4, there's a sharp decrease in harmfulness scores and rates, as well as the AdvBench ASR. This indicates that moderate intervention effectively enhances model safety. However, beyond $\alpha = 4$, we see a gradual increase in these metrics, suggesting that excessive intervention may lead to unintended consequences, potentially disrupting the model's learned safety boundaries. Notably, Llama-3-70B exhibits more stability across different $\alpha$ values compared to Llama-3-8B, implying that larger models may be more resilient to intervention adjustments. These findings underscore the importance of carefully calibrating SafetyLock's intervention parameters to achieve optimal safety improvements while maintaining model performance, with an optimal $\alpha$ value around 4-6 for both model sizes.

\begin{figure}[h]
\begin{center}
	\includegraphics[width=\textwidth]{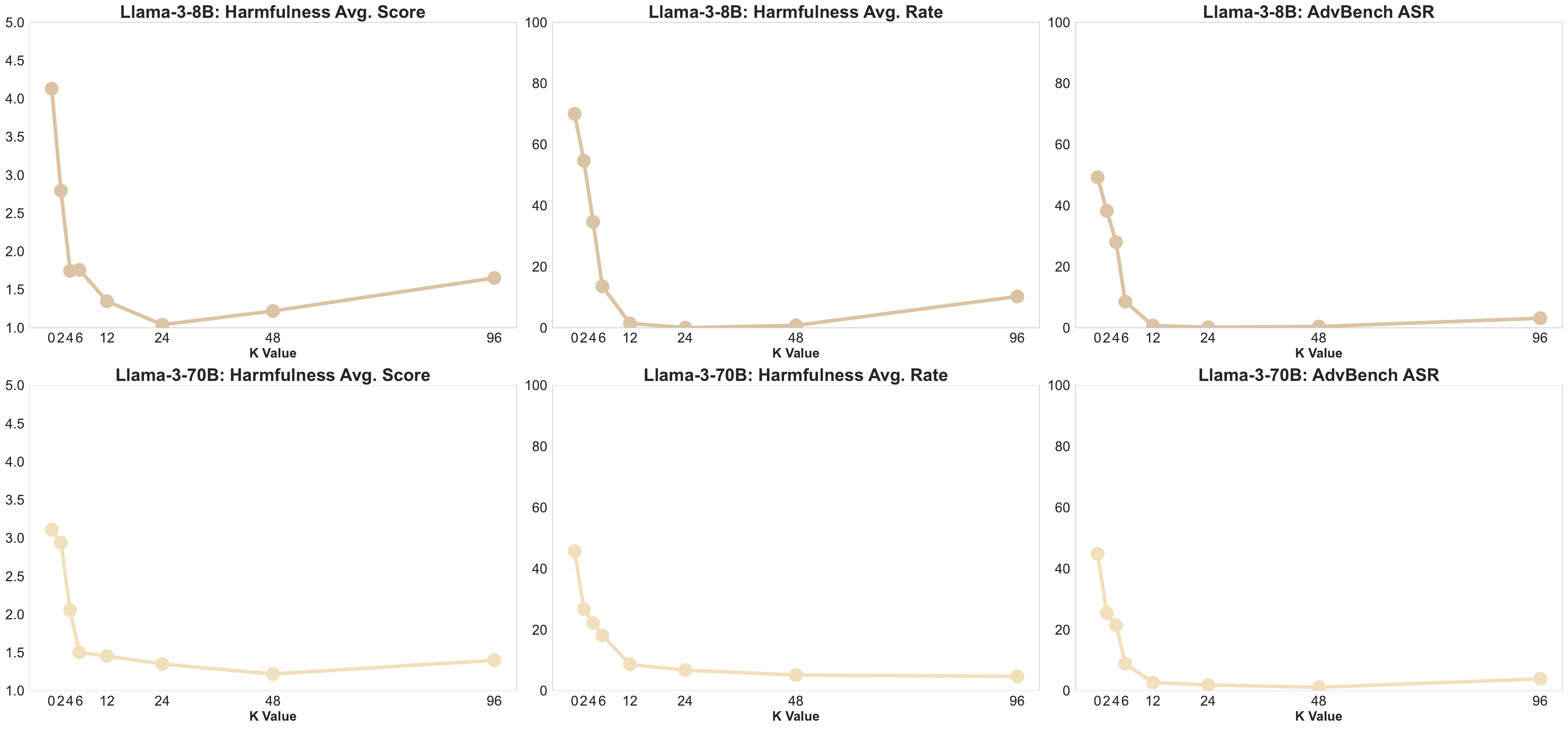}
\end{center}
\caption{Impact of SafetyLock's intervention degree (K) on model safety metrics for Llama-3-8B and Llama-3-70B models. The graphs illustrate the Harmfulness Average Score, Harmfulness Average Rate, and AdvBench ASR across different K values, ranging from 0 to 96. Lower scores indicate better safety performance. Note the rapid improvement in safety metrics as K increases from 0 to 6, followed by more gradual enhancements up to K=24, with a slight uptick at K=96 for some metrics.}
\label{fig:k_impact}

\end{figure}

\textbf{Degree $K$.} Our experiments, as illustrated in Figure \ref{fig:k_impact}, reveal the crucial role of SafetyLock's intervention degree (K) in enhancing model safety across different model sizes. For both Llama-3-8B and Llama-3-70B, we observe a rapid improvement in safety metrics as K increases from 0 to 6, followed by a more gradual enhancement up to K=24. This trend is consistent across all three metrics: Harmfulness Average Score, Harmfulness Average Rate, and AdvBench ASR. For Llama-3-8B, the most significant improvements occur between K=0 and K=6, with the Harmfulness Average Score dropping from about 4.0 to 1.7, and the Harmfulness Average Rate decreasing from 70\% to around 15\%. The AdvBench ASR shows a similar sharp decline. Beyond K=6, the improvements become more incremental, with optimal performance generally achieved around K=24.

Llama-3-70B exhibits a similar pattern but with overall lower harmfulness scores and rates. The initial drop in harmful metrics is less dramatic, suggesting that larger models may have inherently better safety characteristics. However, the trend of improvement with increasing K values remains consistent. Interestingly, for both model sizes, there's a slight uptick in harmfulness metrics for very high K values (K=96), particularly noticeable in the Llama-3-8B model. This suggests that excessive intervention might slightly degrade the model's learned safety boundaries, emphasizing the importance of finding an optimal K value. These findings underscore the effectiveness of SafetyLock in improving model safety, with the most significant gains achieved at relatively low K values (6-24). This implies that targeted intervention on a subset of attention heads can yield substantial safety improvements without the need for exhaustive modification of the model architecture.

\end{document}